\def\cvprPaperID{0000}
\def\confName{CVPR}
\def\confYear{2023}

\def\paperTitle{Variation-Aware Semantic Image Synthesis}

\newif\ifreview 
\newif\ifarxiv 
\newif\ifcamera \newcommand{\cameraready}{\cameratrue}
\newif\ifrebuttal 

\cameraready % \review OR \arxiv OR \cameraready

\pdfoutput=1
\documentclass[10pt,twocolumn,letterpaper]{article}
\ifreview \usepackage[review]{cvpr} \fi
\ifarxiv \usepackage[pagenumbers]{cvpr} \fi
\ifrebuttal \usepackage[rebuttal]{cvpr} \fi
\ifcamera \usepackage{cvpr} \fi

\usepackage{graphicx}
\usepackage{amsmath}
\usepackage{amssymb}
\usepackage{booktabs}

\usepackage{times}
\usepackage{microtype}
\usepackage{epsfig}
\usepackage[table,xcdraw]{xcolor}
\usepackage{caption}
\usepackage{float}
\usepackage{placeins}
\usepackage{color, colortbl}
\usepackage{stfloats}
\usepackage{enumitem}
\usepackage{tabularx}
\usepackage{xstring}
\usepackage{multirow}
\usepackage{xspace}
\usepackage{url}
\usepackage{subcaption}
\usepackage{xcolor}
\usepackage[hang,flushmargin]{footmisc}

\ifcamera \usepackage[accsupp]{axessibility} \fi

% ...

\ifarxiv  \fi

\newcommand{\R}[1]{{%
    \textbf{%
        \ifstrequal{#1}{1}{\textcolor{red}{R#1}}{%
        \ifstrequal{#1}{2}{\textcolor{blue}{R#1}}{%
        \ifstrequal{#1}{3}{\textcolor{magenta}{R#1}}{%
        \ifstrequal{#1}{4}{\textcolor{teal}{R#1}}{%
                           \textcolor{cyan}{R#1}%
        }}}}%
    }%
}}

\usepackage{xr-hyper}

\makeatletter
\newcommand*{\addFileDependency}[1]{
  \typeout{(#1)}
  \@addtofilelist{#1}
  \IfFileExists{#1}{}{\typeout{No file #1.}}
}

\makeatother

\usepackage[pagebackref,breaklinks,colorlinks]{hyperref}
\usepackage[capitalize]{cleveref}
\crefname{section}{Sec.}{Secs.}
\crefname{table}{Table}{Tables}
\crefname{figure}{Fig.}{Figs.}

\begin{document}
%% TITLE
\title{\paperTitle}
% \author{\authorBlock}
\author{Mingle Xu\\
Department of Electronics Engineering, \\
Core Research Institute of Intelligent Robots, Jeonbuk National University, South Korea 
% {\tt\small xml@jbnu.ac.kr}
\and
Jaehwan Lee\\
Department of Electronics Engineering,\\
Core Research Institute of Intelligent Robots, Jeonbuk National University, South Korea 
% {\tt\small dlwo6@jbnu.ac.kr}
\and
Sook Yoon\\
Department of Computer Engineering, Mokpo National University, South Korea 
% {\tt\small syoon@mokpo.ac.kr}
\and
Hyongsuk Kim\\
Department of Electronics Engineering, \\
Core Research Institute of Intelligent Robots, Jeonbuk National University, South Korea 
% {\tt\small hskim@jbnu.ac.kr}
\and
Dong Sun Park\\
Department of Electronics Engineering, \\
Core Research Institute of Intelligent Robots, Jeonbuk National University, South Korea
% {\tt\small dspark@jbnu.ac.kr}
}
\maketitle

\begin{abstract}
Semantic image synthesis (SIS) aims to produce photorealistic images aligning to given conditional semantic layout and has witnessed a significant improvement in recent years. Although the diversity in image-level has been discussed heavily, class-level mode collapse widely exists in current algorithms. Therefore, we declare a new requirement for SIS to achieve more photorealistic images, variation-aware, which consists of inter- and intra-class variation. The inter-class variation is the diversity between different semantic classes while the intra-class variation stresses the diversity inside one class. Through analysis, we find that current algorithms elusively embrace the inter-class variation but the intra-class variation is still not enough. Further, we introduce two simple methods to achieve variation-aware semantic image synthesis (VASIS) with a higher intra-class variation, semantic noise and position code. We combine our method with several state-of-the-art algorithms and the experimental result shows that our models generate more natural images and achieves slightly better FIDs and/or mIoUs than the counterparts. Our codes and models will be publicly available.
\end{abstract}
\section{Introduction}
\label{sec:intro}
Image synthesis, aiming to generate photorealistic images, has been received an improvement in recent years with generative adversarial network (GAN) \cite{goodfellow2014generative, mirza2014conditional}. Seminal works produce images from random noise \cite{goodfellow2014generative, radford2015unsupervised}, while current works produce images from given conditions, such as labels \cite{brock2018large, karras2019style}, words \cite{zhang2017stackgan, zhang2018stackgan++} and images \cite{isola2017image, zhu2017unpaired, wang2018high}. In this paper, we focus on a special condition, semantic image synthesis (SIS) that aims to synthesize images from conditional semantic layouts \cite{park2019semantic, wang2018high, tan2021efficient}.

\begin{figure*}[h!]
\begin{center}
\includegraphics[width=0.99\linewidth]{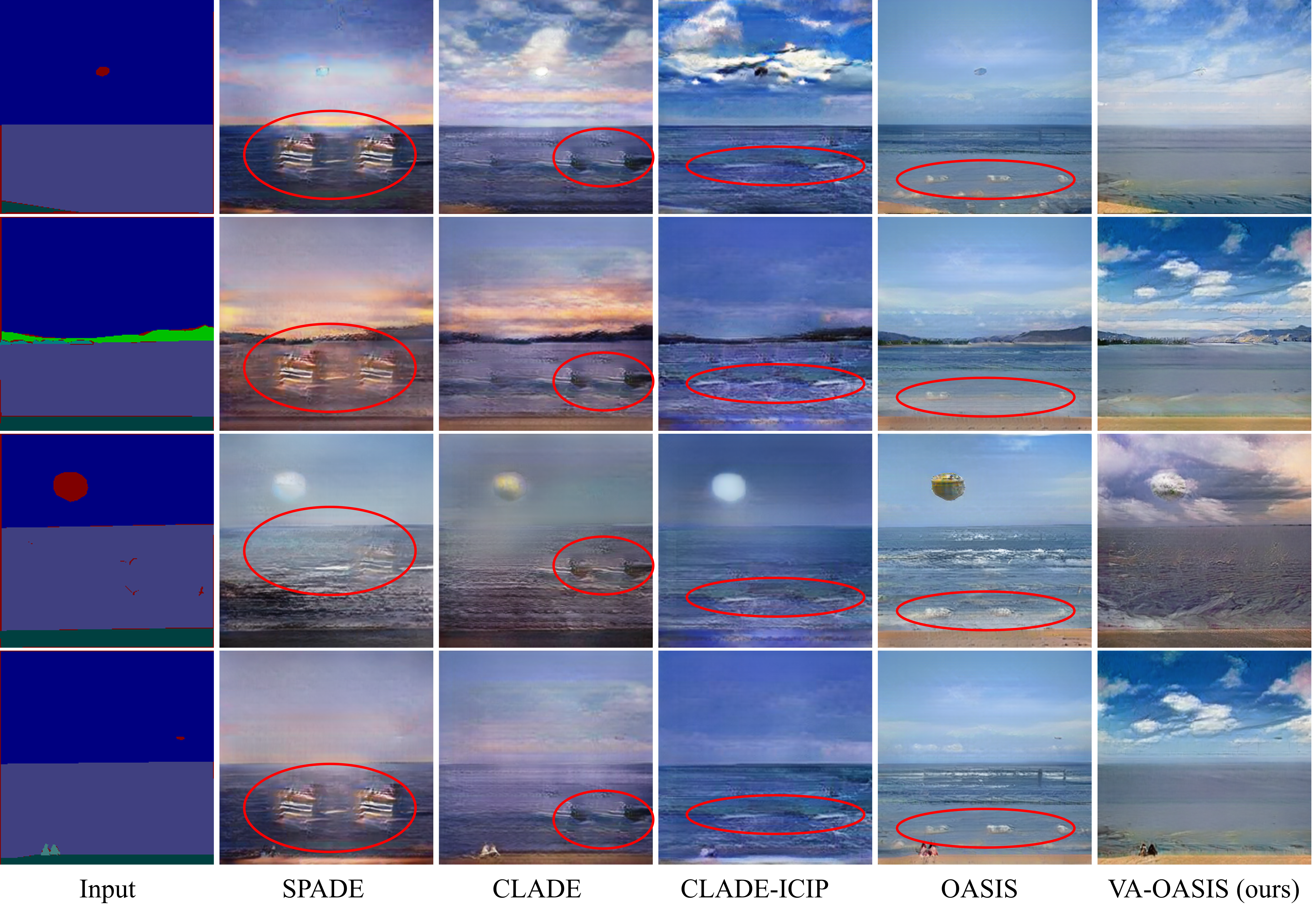}
\end{center}
\caption{Examples of class-level mode collapse on ADE20k. Similar patterns in red circle exist when similar semantic layout are given while our method eases the class-level mode collapse.}
\label{fig1}
\end{figure*}

Mode collapse is one notorious challenge for all image generation applications with GAN and refers to producing similar output \cite{goodfellow2014generative, salimans2016improved, chen2016infogan, isola2017image, zhu2017toward, huang2018multimodal, lee2018diverse, park2019semantic, zhu2020semantically, tan2022semantic}. For example, in image to image translation, one desired thing is producing multiple possible images with a same conditional image \cite{chen2016infogan, zhu2017toward, huang2018multimodal, lee2018diverse}, in which the challenge is formally called multimodal image synthesis. Similarly, a single semantic layout can align with multiple photorealistic images \cite{park2019semantic, zhu2020semantically, tan2022semantic}. Although multimodal SIS has been heavily discussed in recent years, most them consider it in \emph{image-level} where one image is different from another image and the diversity in \emph{class-level} is somehow ignored and underdeveloped. The class-level diversity is inspired by the fact that semantic layouts give more details than conditional object labels in image-level, where the pixels belonging to a same semantic class may not always have same pixels in generated images. Therefore, mode collapse can be analogical from image-level to class-level. As shown in Figure \ref{fig1}, class-level mode collapse can be observed in three current state-of-the-art algorithms. The generated images may be different from others, whereas similar patterns appear when similar semantic layout are given.

To analyze the mode collapse challenge in SIS, we consider the diversity not only in image-level but also in class-level and hence, we define \emph{inter-class} and \emph{intra-class} variation. The inter-class variation is the diversity between different semantic classes while the intra-class variation stresses the diversity inside one class. Furthermore, we probe three current state-of-the-art algorithms, SPADE \cite{park2019semantic}, CLADE \cite{tan2021efficient}, and OASIS \cite{schonfeld2020you, sushko2022oasis}. Through analysis, we find that the conditional normalization in the three algorithms elusively contributes to inter-class variation whereas the intra-class variation is somehow ignored and not enough. The details are discussed in Section \ref{analysis}. Besides, the less intra-class variation results in the mode collapse in class-level, as shown in Figure \ref{fig1}. Through our analysis, we declare a new requirement to ease the two-type mode collapse in SIS, \emph{variation-aware}, consisting of inter- and intra-class variation. We term this goal variation-aware semantic image synthesis (VASIS) where the generated images should have not only inter-class variation but also intra-class variation.

Holding the variation-aware requirement and to achieve VASIS, we introduce two simple methods to enlarge the intra-class variation, semantic noise and position code. Simultaneously, we argue that intra-class variation is entailed to produce realistic images for SIS. On one hand, semantic noise is Gaussian distribution with learnable shift and scale. As learned individually for each semantic class, we call it semantic noise. Distinguished from the shared noise for each semantic class \cite{brock2018large, karras2019style, schonfeld2020you, sushko2022oasis}, semantic noise can improve the intra-class variation and maintain the inter-class variation. On the other hand, position code enables image generator heterogeneous regarding to the pixel position. When same semantic class exists in different position, the generated pixels may be diverse. Specially, two more types of position codes are discussed, except for relative position code \cite{tan2021efficient} and the learnable position code achieves the best among them. We verify our ideas and the proposed two methods in three current state-of-the-art and the experimental result shows that our models generate more natural images and achieves slightly better FIDs and/or mIoUs than the counterparts. Our codes and models will be publicly available.
\section{Related Work}
\label{sec:related}

\textbf{Semantic image synthesis,} aiming to generate photorealistic images aligning the given semantic layout, is the main objective of this paper. The first challenge of this task is how to leverage the given semantic layout effectively and efficiently. To address this challenge, preliminary works adopt an encoder to extract the input semantic layout, followed by a decoder to produce images \cite{isola2017image, zhu2017unpaired, wang2018high, tang2020dual}. However, this encoder-decoder strategy is declared to "wash away" the input semantic layout \cite{park2019semantic} and simultaneously, SPADE is proposed to utilize the input semantic layout more effectively. In SPADE \cite{park2019semantic}, the encoder is discarded and the given semantic layout is directly integrated into a decoder via a conditional normalization layer. Following SPADE, the decoder with a conditional normalization layer is widely employed in many other papers for different goals, such as enabling every instance with specific style \cite{zhu2020sean}, high resolution semantic layout synthesis \cite{li2021collaging}, semantic multi-modal image synthesis \cite{zhu2020semantically, tan2021diverse}, improving the synthesis quality \cite{schonfeld2020you}, and efficient semantic image synthesis \cite{tan2021efficient, xu2022enhanced}. In this paper, we focus on how to achieve the variation-aware semantic image synthesis which requires that each class should be different from any other classes (inter-class variation) and each class owns its diversity (intra-class variation).

\textbf{Normalization layer} is firstly designed to accelerate the training process by reducing the internal covariate Shift \cite{ioffe2015batch}. The batch normalization layer consists of two steps, normalization with computed mean and variance and de-normalization with learnable shift and scale, both of which are adapted afterward. On one hand, the way to compute the mean and variance changes from a batch, such as instance normalization \cite{ulyanov2016instance}, group normalization \cite{wu2018group}, layer normalization \cite{ba2016layer}. On the other hand, the learnable shift and scale become conditional version, such as style-conditional \cite{huang2017arbitrary},  label-conditional \cite{brock2018large, karras2019style} and semantic layout-conditional \cite{park2019semantic, tan2021efficient}. In this paper, the conditional normalization is borrowed to achieve a new objective, variation-aware semantic image synthesis, with extra semantic noise and position code.

\textbf{Image variation} is heavily analyzed in image classification with convolution neural networks (CNN) \cite{zeiler2014visualizing, karras2019style} and other work \cite{xu2022comprehensive}. Image classification and other dense prediction ask model learn the inter-class variation via recognizing and overlooking the intra-class variations. But image generation and semantic image synthesis are required to synthesize intra-class variation. However, the diversity in class-level is somehow ignored and underdeveloped, though image-level diversity is a hot topic \cite{karras2019style, brock2018large, lee2018diverse, zhu2017toward}. In this paper, we introduce two simple methods to enlarge class-level diversity, semantic noise and position code.

\textbf{Position code.} The convolution neural network (CNN), playing an important role in the era of deep learning, owns an inductive bias because of the local connection and shared weight, with which the position of a pattern is independent of the feature extracting process. However, the position-dependent sometimes maybe the desired character. By considering the issue, the absolute position map is concatenated with the input feature map before undergoing the CNN operation \cite{liu2018intriguing}. Moreover, from the CNN to transformers, the position code becomes more popular and is adopted in many research work \cite{vaswani2017attention, dosovitskiy2020image, wu2021rethinking, gehring2017convolutional, shaw2018self}. Different from the CNNs, the position codes in transformers are utilized to mimic the word sequence from the natural language processing \cite{devlin2018bert}. The absolute position code is originally borrowed \cite{vaswani2017attention} while the learnable absolute position code \cite{gehring2017convolutional} shows its convenience with the changeable length and leads to better performance. Afterward, the relative position code is advocated as it captures the connections between two elements. Although it is introduced in SIS \cite{tan2021efficient}, position code was not systematic to be analyzed. In this paper, we discuss three types of position code to improve the intra-class variation when performing semantic image synthesis and ease the class-level mode collapse.

\section{Method}
\label{sec:method}

We observe and aim to ease the class-level mode collapse in semantic image synthesis (SIS). In this section, we first analyze how the current methods produce inter-class variation, the diversity between semantic classes, and intra-class variation, the diversity inside one specific semantic class. Then we propose two simple methods to enlarge the intra-class variation and thus ease the class-level mode collapse.

\subsection{Revisit SPADE and CLADE}
\label{analysis}

Let $s=S^{N \times H \times W}$ be the input semantic image, where $N$ is the number of semantic labels, $H$ and $W$ are the height and width. Given specific location $h,w$, each entry $s_{:,h,w}$ denotes the semantic label for the specific location in a one-hot label manner. The spatial-adaptively normalization (SPADE) can be formalized as

\begin{equation}
    \hat{x}_{i,j,k} = \gamma_{i,j,k} \times \frac{x_{i,j,k}-\mu_{i,j,k}(x)}{\sigma_{i,j,k}(x)} + \beta_{i,j,k},
\end{equation}
with the coordinate channel $i$, height $j$, width $k$. $x$ is the input feature and $\hat{x}$ denotes the output feature map. $\mu(x)$ and $\sigma(x)$ are the mean and variance of the input feature. Different from unconditional normalization \cite{ioffe2015batch, ulyanov2016instance, ba2016layer, wu2018group}, the scale $\gamma$ and shift $\beta$ in SPADE \cite{park2019semantic} are computed from semantic layout $s$. Specifically, $\gamma = \mathcal{F}_2(\mathcal{F}_1(s))$ and $\beta = \mathcal{F}_3(\mathcal{F}_1(s))$ where $\mathcal{F}$ is a convolution neural network (CNN) with three as their kernel sizes. As CNN is locally connected and weight shared, the same semantic label results in the same scale and shift vector \cite{xu2022enhanced}. Mathematically, given two semantic labels $s_1 =s_{:,j_1,k_1}$ and $s_2=s_{:,j_2,k_2}$ along with the vicinity semantic label $s_{\mathcal{V}_{s1}}$ and $s_{\mathcal{V}_{s2}}$, if $s_1=s_2=s_{\mathcal{V}_{s1}}=s_{\mathcal{V}_{s2}}$, then $\gamma_{:,j_1,k_1}=\gamma_{:,j_2,k_2}$ and $\beta_{:,j_1,k_1}=\beta_{:,j_2,k_2}$. More efficiently, CLADE \cite{tan2021efficient} replaces the three CNN layers in SPADE with a sampling strategy, which leads to much less number of parameters and computations (FLOPs). Hence, $\gamma_{:,j_1,k_1}=\gamma_{:,j_2,k_2}$ and $\beta_{:,j_1,k_1}=\beta_{:,j_2,k_2}$ are also satisfied for all $s_1=s_2$ as CLADE does not fuse pixel with its vicinity region. Based on the above analysis, the first question is how SPADE and CLADE can generate different pixel values if each semantic label has the specific shift and scale vector. 

\begin{table}
\begin{center}
\begin{tabular}{|c|c|c|c|c|}
\hline
& & FID-t & mIoU & Acc \\
\hline
\multirow{3}{*}{SPADE} 
& original & 51.98 & 62.21 & 93.48 \\
& reflect & 49.13 & 61.32 & 93.43 \\
& conv-k1 & 285.18 & 8.44 & 25.09 \\
\hline
\multirow{3}{*}{CLADE}
& original & 50.62 & 60.63 & 93.50 \\
& reflect & 49.98 & 60.07 & 93.33 \\
& conv-k1 & 255.55 & 9.86 & 29.82 \\
\hline
\end{tabular}
\end{center}
\caption{Understanding the intra-class variation in SPADE and CLADE. We probe the zero-padding and the kernel size of CNN by replacing the zero-padding with the reflect-padding (reflect) and the kernel size three with one (conv-k1).}
\label{table1}
\end{table}

\begin{figure}[h!]
\begin{center}
\includegraphics[width=1.0\linewidth]{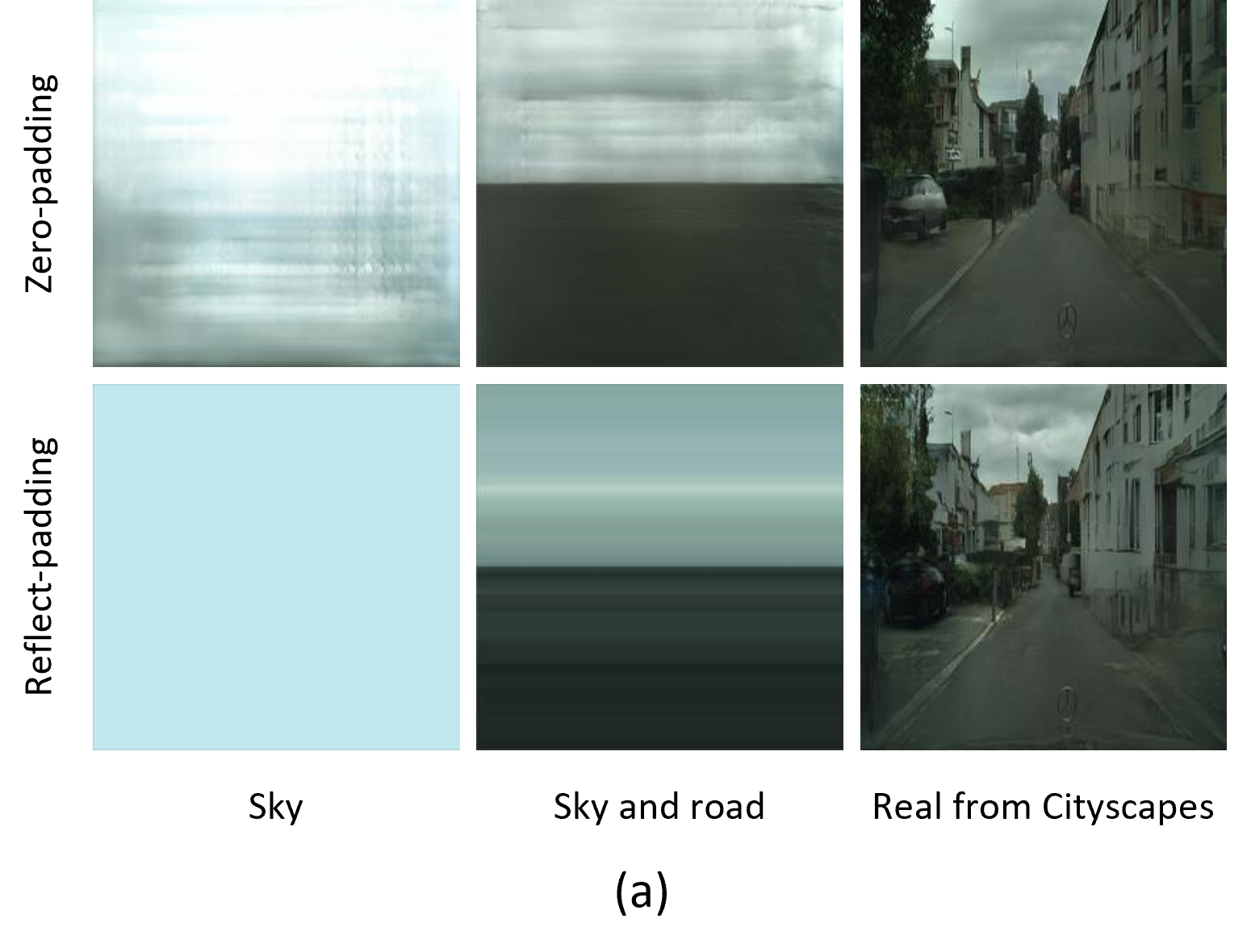} \\
\includegraphics[width=1.0\linewidth]{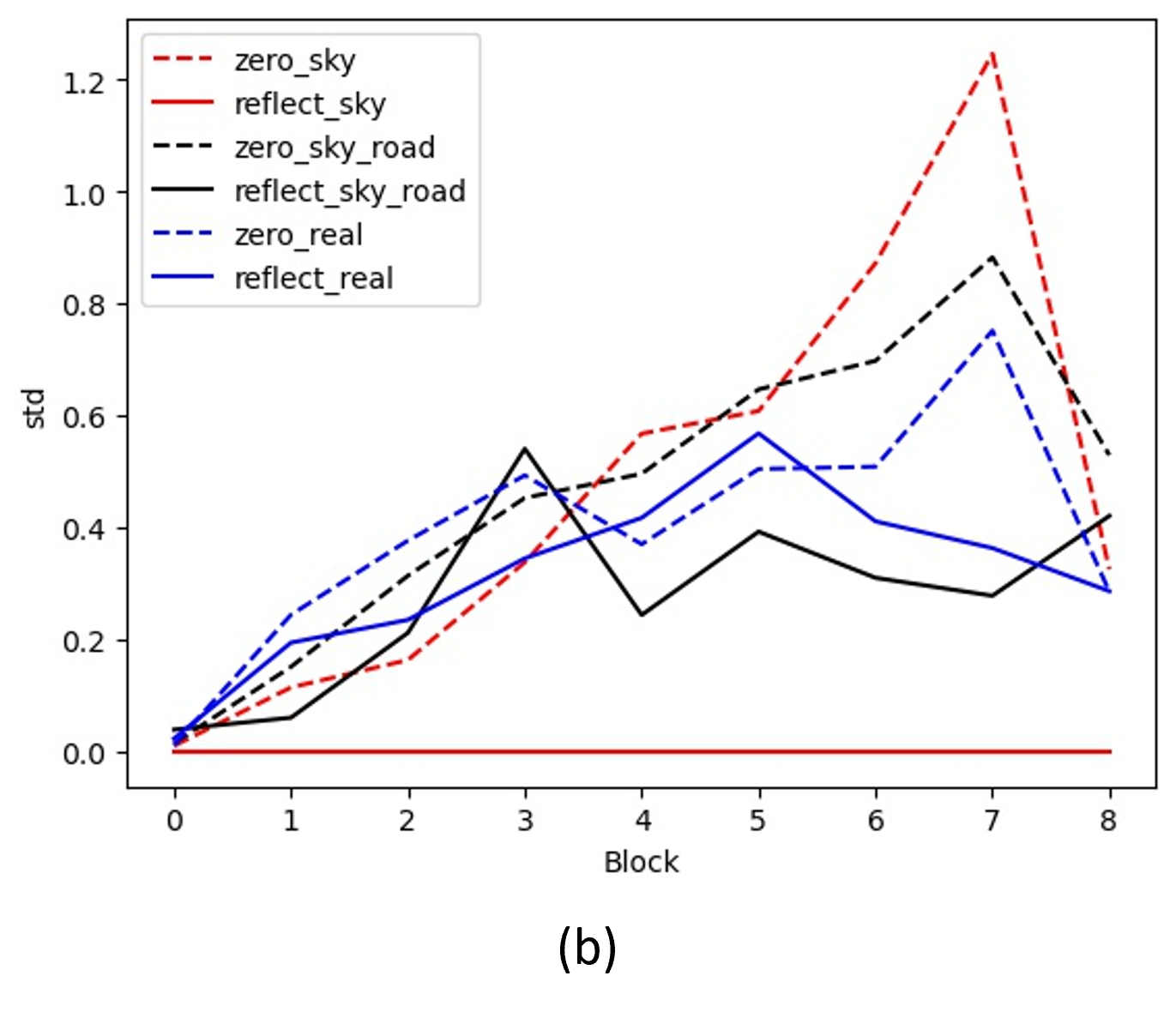}
\end{center}
\caption{(a) The generated images by replacing the zero-padding with the reflect-padding with model trained in Cityscapes dataset. (b) The standard deviation of features in different blocks.}
\label{fig3}
\end{figure}

A quick answer is the coefficient of class boundary and conv-k3 (three as the kernel size of convolution layer). Let us define \emph{class boundary} in semantic layout, $s_1 \neq s_{\mathcal{V}_{s1}}$ that are from zero padding and input semantic layout. Conv-k3 fuses semantic labels on the class boundary, which makes different shift and scale vectors for each semantic label. To probe conv-k3 impact, we replace them with conv-k1 (one as the kernel size). As shown in Table \ref{table1}, both SPADE and CLADE do not work, with unreasonable FID and mIoU. Although conv-k3 contributes to intra-class variation, it occurs must with the class boundary. Similar class boundaries result in similar patterns, which is termed as class-level mode collapse in this paper. However this mode collapse is not easy to observe due to two factors. First, natural semantic layout in current datasets is taken as the input that may with multiple class boundary and the class boundaries are heterogeneous in different images. As shown in Figure \ref{fig3}, the generated images from Cityscapes dataset seems reasonable because of diverse class boundaries in the input semantic layout. But when similar class boundary appears as shown in Figure \ref{fig1}, the mode collapse dominates even that the class boundary appears in diverse locations. 

Second, zero padding also leads to class boundary. As shown in the top row in Figure \ref{fig3}, the generated images with one or two semantic class in one image do not clearly show the mode collapse. But replacing the zero padding with reflect padding, mode collapse is amplified, such as the horizontal patterns in the sky and road case because of horizontal class boundary. Surprised, reflect padding benefits FID and lessen mIoU for both SPADE and CLADE as displayed in Table \ref{table1}. To have a deeper understanding, we probe the standard deviation of feature maps in different blocks. As shown in Figure \ref{fig3}, reflect padding with only sky class gives no variation in all blocks, while zero padding make a fake variations. The deviation can be not distinguished when with multiple semantic classes and zero padding, which verifies that class-level mode collapse is elusive to observe. Besides, the deviation is improved in the early several blocks as the class boundary is increasing.

To conclude, the conditional normalization layer in SPADE \cite{park2019semantic}, CLADE \cite{tan2021efficient}, and other similar papers \cite{zhu2020sean, schonfeld2020you, sushko2022oasis} leads to inter-class variation via giving every semantic class a specific shift and scale vector. Simultaneously, the intra-class variation results from class boundary and conv-k3 but suffer mode collapse.

\textbf{Discussion.} Our analysis supports previous findings. For example, SPADE has a slight superiority with conv-k3 than conv-k1 inside the conditional normalization layer because of chances to have more class boundaries. CLADE is inferior to SPADE as the class boundary is not used in its conditional normalization layer and CLADE-ICPE is bettern than plain CLADE. Noise is beneficial to OASIS \cite{schonfeld2020you, sushko2022oasis}. Among them, better one tends to have an improved intra-class variation. Besides, intra- and inter-class variation can explain the inferior FID in \cite{shaham2021spatially} since that the shared weight in a local area results in lower inter- and intra-class variation.

\subsection{Variation-aware semantic image synthesis}

Through the above analysis, we argue that the intra-class variation is not enough in the current algorithms, which results in undesired patterns and artifacts. To address this issue, we propose a requirement that a model should own intra-class and inter-class variation simultaneously to synthesize more natural images, termed variation-aware semantic image synthesis (VASIS). Further, we propose a simple VASIS model by adding semantic noise and position code into the original conditional normalization layer. As the conditional normalization layer enables the same semantic label with the same shift or scale vector, our motivation is to introduce randomness into the vectors. In another word, we aim to make the shift and scale for each semantic label vary in a distribution. 
% We argue that this challenge is not trivial as we should achieve a good intra-class variation without reducing the inter-class variation. 
Given a feature map $x\in \mathbb{R}^{B \times C' \times H' \times W'}$ with batch size $B$, channel $C'$, height $H'$, and width $W'$, the scale in our VASIS model is formalized as (the shift is computed in the same way and omitted):

\begin{equation}
    \gamma = \gamma_{n} \bigoplus (\gamma_s \bigotimes \gamma_p),
\end{equation}
where $\gamma \in \mathbb{R}^{B \times C' \times H' \times W'}$. $\bigoplus$ and $\bigotimes$ denote the concatenation in channel-wise and the multiplication in element-wise, respectively. Our $\gamma$ consists of three parts, semantic noise $\gamma_n \in \mathbb{R}^{B \times \frac{C'}{2} \times H' \times W'}$, semantic-layout feature $\gamma_s \in \mathbb{R}^{B \times \frac{C'}{2} \times H' \times W'}$ and position code $\gamma_p \in \mathbb{R}^{B \times \frac{C'}{2} \times H' \times W'}$.
The semantic layout feature is computed same as the original method, such as convolution layer in SPADE or sampling strategy in CLADE.

\textbf{Semantic noise} is calculated as follows:
\begin{equation}
    \gamma_n = \mathcal{N} \bigotimes \mathcal{S}(s, \mathcal{N}_1) + \mathcal{S}(s, \mathcal{N}_2),
\end{equation}
where $\mathcal{N}\in \mathbb{R}^{B \times \frac{C'}{2} \times H' \times W'}$ denotes a standard normal distribution. $\mathcal{N}_1 \in \mathbb{R}^{N \times \frac{C'}{2}}$ and $\mathcal{N}_2 \in \mathbb{R}^{N \times \frac{C'}{2}}$ are learnable parameters. $\mathcal{S}$ suggests the same guided sampling strategy in CLADE \cite{tan2021efficient} and CSN \cite{xu2022enhanced} that samples a vector from $\mathcal{N}_1$ or $\mathcal{N}_2$ according to the semantic label $s\in \mathbb{R}^{B \times N \times H' \times W'}$ where $N$ is the number of semantic labels. Our semantic noise distinguishes from random noise as utilized in \cite{schonfeld2020you}. If the random noise are borrowed to compute the shift and scale for all classes, then the intra-class variation is eased but the inter-class variation is reduced as all classes shares one random noise distribution. In contrast, the semantic noise mitigates the former one without sacrificing the latter one. We emphasis that reducing the number of channels of semantic-layout feature, from $C'$ to $\frac{C'}{2}$, contributes to smaller number of parameters for SPADE and OASIS but seems no large impact on the performance.

\textbf{Position code} is formalized as follows:

\begin{equation}
    \gamma_p = \mathcal{F}_p(p_l),
\end{equation}
where $\mathcal{F}_p$ is a position code function, a convolution layer. $p_l\in \mathbb{R}^{1 \times 2 \times H' \times W'}$ denotes learnable position code. Except for $p_l$, other position codes are also verified, such as absolute position code $p_a$ \cite{liu2018intriguing} and relative position code $p_r$ \cite{tan2021efficient}. To be specific, $p_a$ calculates the position code according to the absolute location of each pixel in height-wise and width-wise.
The relative position encoded the distance between one pixel and the center of its same semantic area. Table \ref{table2} lists the differences of their characters. We consider the position code in two ways. The first one is the computation and we find that the relative one needs computation, averagely 0.32 second per image in Cityscapes dataset. Second, is the value monotonic along a direction? The absolute and the relative ones are monotonic, which may result in a gradually changing pattern in the generated images. But the learnable one is not monotonic as it is learned via the optimization method.

\textbf{Discussion.}
We introduce two simple methods to improve the intra-class variation with minor changes, but some other alternatives can be considered. One may wonder the situation to replace noise with the input semantic layout in SPADE and CLADE. We argue that the input noise may be ignored by the model mainly because of the perception loss as mentioned in pix2pix \cite{isola2017image}. Our preliminary verified the assumption. Besides, more loss function can be utilized, such VAE loss \cite{kingma2013auto, park2019semantic, liu2019learning} and mode seeking loss \cite{mao2019mode}. A systematic analysis is leaved as our future work. Instead, we aim to probe the class-level mode collapse in semantic image synthesis and adapt randomness into the architecture to ease the issue.

\begin{table}[h!]
\begin{center}
\begin{tabular}{|c|c|c|c|}
\hline
& $p_a$ & $p_l$ & $p_r$ \\
\hline
no computation & yes & yes & no \\
\hline
not monotonity & no & yes  & no \\
\hline
% image-based & $\times$ & $\times$ & yes \\
% \hline
\end{tabular}
\end{center}
\caption{Characters of the three position-codes, considered in two ways. Do we need to compute them and is the value monotonic along a direction?}
\label{table2}
\end{table}

\section{Experiments}

\subsection{Experimental settings}

\begin{table*}[h!]
\begin{center}
\begin{tabular}{|c|c|c|c|c|c|c|c|c|c|c|}
\hline
\multirow{2}{*}{Dataset} & \multirow{2}{*}{Method} & \multirow{2}{*}{FID-t$\downarrow$} & \multirow{2}{*}{FID-v$\downarrow$} & \multirow{2}{*}{mIoU$\uparrow$} & \multirow{2}{*}{Acc$\uparrow$} & \multirow{2}{*}{b/n} & \multicolumn{2}{c|}{Generator} & \multicolumn{2}{c|}{Discriminator} \\
\cline{8-11} & & & & & & & para & FLOPs & para & FLOPs \\
\hline
\multirow{13}{*}{Cityscapes}
& LGGAN & 52.52 & 59.17 & 66.55 & 94.05 & 8/8 & 111.12 & 475.73 & 5.60 & 10.48 \\
& CC-FPSE & 43.80 & 50.43 & 65.33 & 93.91 & 32/16 & 128.06 & 739.30 & 5.18 & 6.65 \\
\cline{2-11}
& SPADE &51.98 & 58.71 & 62.18 & 93.47 & 16/8 & 93.05 & 281.64 & 5.60 & 10.48 \\
& SPADE* & 49.17 & 55.83 & 62.50 & 93.60 & 16/4 & 93.05 & 281.64 & 5.60 & 10.48 \\
& VA-SPADE & 46.28 & 53.30 & 61.08 & 93.40 & 16/4 & 83.34 & 189.61 & 5.60 & 10.48 \\
\cline{2-11}
& CLADE & 50.62 & 56..69 & 60.69 & 93.50 & 16/8 & 67.90 & 75.58 & 5.60 & 10.48 \\
& CLADE* & 54.82 & 62.28 & 56.43 & 92.95 & 16/4 & 67.90 & 75.58 & 5.60 & 10.48 \\
& CLADE-ICPE & 42.38 & 50.50 & 60.71 & 93.35 & 16/8 & 67.90 & 75.58 & 5.60 & 10.48 \\
& CLADE-ICPE* & 54.53 & 61.13 & 58.88 & 93.17 & 16/4 & 67.90 & 75.58 & 5.60 & 10.48 \\
& VA-CLADE & 48.14 & 55.18 & 62.25 & 93.45 & 16/4 & 70.41 & 75.82 & 5.60 & 10.48 \\
\cline{2-11} 
& OASIS & 41.71 & 48.37 & 67.01 & 92.61 & 20/4 & 71.11 & 319.03 & 22.25 & 97.05 \\
& OASIS* & 43.42 & 49.25 & 68.42 & 92.81 & 16/4 & 71.11 & 319.03 & 22.25 & 97.05 \\
& VA-OASIS & 41.39 & 48.63 & 67.78 & 92.97 & 16/4 & 61.82 & 183.33 & 22.25 & 97.05 \\
\hline

\multirow{13}{*}{ADE20k} 
& LGGAN & 27.60 & 35.23 & 38.95 & 79.63 & 24/8 & 115.09 & 312.57 & 5.84 & 7.83 \\
& CC-FPSE & 29.88 & 37.30 & 40.49 & 80.75 & 32/16 & 140.70 & 438.25 & 5.19 & 4.29 \\
\cline{2-11}
& SPADE & 29.80 & 37.51 & 38.27 & 79.22 & 32/8 & 96.49 & 181.33 & 5.84 & 7.83 \\
& SPADE* & 29.31 & 37.28 & 37.19 & 78.98 & 32/4 & 96.49 & 181.33 & 5.84 & 7.83 \\
& VA-SPADE & 28.77 & 37.10 & 36.76 & 77.98 & 32/4 & 88.26 & 134.75 & 5.84 & 7.83 \\
\cline{2-11}
& CLADE & 30.50 & 38.23 & 37.01 & 78.40 & 32/8 & 71.43 & 42.25 & 5.84 & 7.83 \\
& CLADE* & 29.93 & 37.80 & 36.26 & 78.40 & 32/4 & 71.43 & 42.25 & 5.84 & 7.83 \\
& CLADE-ICPE & 28.70 & 36.77 & 36.59 & 78.02 & 32/8 & 71.43 & 42.25 & 5.84 & 7.83 \\
& CLADE-ICPE* & 28.75 & 36.64 & 36.79 & 78.44 & 32/4 & 71.43 & 42.25 & 5.84 & 7.83 \\
& VA-CLADE & 29.16 & 36.70 & 36.90 & 78.68 & 32/4 & 74.16 & 42.37 & 5.84 & 7.83 \\
\cline{2-11} 
& OASIS & 27.49 & 35.39 & 45.62 & 82.93 & 32/4 & 74.31 & 194.56 & 22.26 & 49.02 \\
& OASIS* & 28.52 & 36.45 & 43.90 & 82.05 & 24/4 & 74.31 & 194.56 & 22.26 & 49.02 \\
& VA-OASIS & 27.43 & 35.37 & 43.36 & 81.45 & 24/4 & 66.05 & 125.93 & 22.26 & 49.02 \\
\hline

\multirow{8}{*}{COCO-Stuff} 
& CC-FPSE & 25.44 & 30.03 & 41.96 & 70.79 & 32/16 & 141.94 & 456.12 & 5.19 & 4.56 \\
\cline{2-11}
& SPADE & 27.70 & 32.75 & 38.15 & 68.86 & 32/8 & 97.48 & 191.32 & 5.90 & 8.54 \\
& VA-SPADE & 27.49 & 32.52 & 37.96 & 68.92 & 32/4 & 89.97 &144.7 & 5.90 & 8.54 \\
\cline{2-11}
& CLADE & 29.17 & 34.47 & 38.16 & 69.18 & 32/8 & 72.51 & 42.43 & 5.90 & 8.54 \\
& CLADE-ICPE & 27.77 & 32.96 & 37.77 & 68.66 & 32/8 & 72.51 &42.43 & 5.90 & 8.54 \\
& VA-CLADE &29.26 & 34.59 & 37.03 & 67.95 & 32/4 & 75.59 & 42.55 & 5.90 & 8.54 \\
\cline{2-11}
& OASIS & 24.69 & 29.64 & 45.28 & 74.32 & 32/8 & 75.20 & 204.23 & 22.26 & 49.16 \\
& VA-OASIS & 24.68 & 29.79 & 43.53 & 73.65 & 24/4 & 67.50 & 135.60 & 22.26 & 49.16 \\
\hline
\end{tabular}
\end{center}
\caption{Comparison with other methods across datasets. We retrain SPADE, CLADE, CLADE-ICPE, and OASIS in our machine with less number of devices to have a fair comparison, which is denoted with *. b/n suggests batch size and number of GPUs in the training process.}
\label{table5}
\end{table*}

\textbf{Dataset.} We execute experiments on three challenging datasets for SIS application, Cityscapes \cite{cordts2016cityscapes}, ADE20k \cite{zhou2017scene}, and COCO-Stuff \cite{caesar2018coco}. Cityscapes owns 2,975 images for training and 500 images for testing with 35 labels related to street scenes, in which instance-level annotation are given. ADE20K includes 20,210 training and 2,000 validation images, with 150 semantic classes but without instance-level annotation. COCO-stuff, with 182 semantic labels and instance-level annotation, is a more challenge dataset having 118,287 training images and 5,000 validation images. Similar to other algorithms \cite{park2019semantic, tan2021diverse, sushko2022oasis}, the original training and testing or validation dataset are directly utilized.

\textbf{Implementation details.} We use the same loss functions with SPADE and CLADE for our version VA-SPADE and VA-CLADE, GAN loss in hinge, perceptual loss, and feature matching loss. Accordingly, our VA-OASIS only uses the GAN loss with a segmentation-based discriminator. Besides, same sizes of images are produced. To be specific, the resolution is $256*512$ for Cityscapes and $256*256$ for ADE20k and COCO-Stuff. Other training receipt are also same, including epochs and learning rate. Differently, we can not access to one sever with eight GPUs, instead four GPUs. Therefore, part of the experiments can not be trained with the same batch size because of the memory issue. To have a fair comparison, we also retrain parts of the compared methods with same number of batch size, given in Table \ref{table5} and emphasised with $*$.

\textbf{Evaluation metrics.} We utilize three metrics to compare different algorithms, FID, mIoU, and Acc. FID compute the distance of distribution between the generated images and the real images. Although it symbols the fidelity of the generated images, FID can not evaluate the matching of the generated images on the conditional semantic layout. Acc and mIoU are generative and detection metric where the generated images is segmented by a trained model and then compute the matching. Even the three metrics are widely used, there are no conforming program and thus it is hard to compare different algorithms. We aim to give a clear description and public codes. To compute FID\footnote{https://github.com/mseitzer/pytorch-fid}, two types of distribution are adopted, the aligned real validation images in the validation dataset (FID-v) and real training images (FID-t). Before computing FID, the real images are down-sampled in Bicubic way to the same resolution as the generated images. To calculate Acc and mIoU, pretrained model are leveraged as mentioned here\footnote{https://github.com/NVlabs/SPADE/issues/39}. To be more specific, pre-trained models from \cite{yu2017dilated}, \cite{zhou2018semantic} and \cite{chen2017deeplab} are utilized to perform the semantic segmentation for synthesized images from Cityscapes, ADE20k, and COCO-Stuff, respectively. Further, during the training process, the best model is recorded based on FID every five epochs for Cityscapes and ten for ADE20k and COCO-Stuff. The evaluation code will be public to encourage clear comparison\footnote{https://github.com/xml94/VASIS}.

\textbf{Compared methods.} SPADE \cite{park2019semantic}, CLADE \cite{tan2021efficient}, and OASIS \cite{schonfeld2020you, sushko2022oasis} are baselines to prove our observation and method. SPADE discards the traditional encoder but leverages a conditional normalization to use the input semantic layout. Further, CLADE finds that sampling from the semantic layout contributes to less parameters and more efficient framework, but slightly shrinks the performance. OASIS employs the same conditional normalization as SPADE but with a segmentation-based discriminator. Figure \ref{fig1} illustrates that all them suffer from the class-level mode collapse. Besides, LGGAN \cite{tang2020local}, adopting the traditional encoder-decoder scheme, is also compared. We also compare with CC-FPSE \cite{liu2019learning} that utilizes a prediction module for CNN weight, instead of shift and scale of normalization layer in SPADE and CLADE. More importantly, VAE loss is employed in CC-FPSE, which may ease the intra-class variation.

\subsection{Main results}

\textbf{Quantitative result.} The main comparison is given in Table \ref{table5}. First, our methods tend to have better FID-t and FID-v but slightly worse mIoU and Acc when training with same number of batch size and GPUs in Cityscapes and ADE20k, which is similar to previous result that adding noise or improving intra-class variation leads to less mIoU and Acc in CLADE \cite{tan2021efficient} and OASIS \cite{schonfeld2020you, sushko2022oasis}. For example, our VA-OASIS achieves FID-t 41.39 and FID-v 48.63 in Cityscapes, lower than 43.42 and 49.25 from OASIS. Except for ADE20k, our VA-CLADE gets worse FID but superior mIoU and Acc than CLADE-ICPE. Besides, our VA-SPADE and VA-OASIS employ much less number of parameters and computations than original SPADE and OASIS but achieve similar or even better results, which suggests that the computations and number of parameters among current algorithms can be reduced.

Second, batch size and number of GPUs non-trivially play impacts on the performance. For instance, SPADE gets all better performance in Cityscapes with less number of GPUs and same batch size as the original paper. However, the trend is not similar across all datasets and models. Especially, CLADE-ICPE receives a much worse performance in Citycapes, such as FID-t 54.53, compared to originally reported 42.38. We guess that the batch normalization layers used in the algorithms credits to this situation. Third, compared to non-encoders architectures, including SPADE, CLADE, OASIS, and our method, other methods may get similar results but with much bigger computation and larger number of parameters. For example, LGGAN and CC-FPSE obtains similar FID-t in ADE20k with SPADE and our method, but with more than double and triple FLOPs.

\textbf{Qualitative result.} Figure \ref{fig1} shows four generated images from ADE20k to prove our assumption and algorithm. The generated images by the current state-of-the-art have similar artifacts or patterns with heterogeneous semantic layout as input. Besides, we observe that the similar patterns exist in same class boundaries. Ours algorithm, VA-OASIS, largely eases the class-level semantic collapse due to semantic noise and position code that introduce diversity for same semantic class, although similar pattern appears, such as the second and last row in Figure \ref{fig1}. More results in Cityscapes and COCO-Stuff refer to supplementary material due to page limitation.

\begin{figure*}[h!]
    \centering
        \includegraphics[width=0.15\textwidth]{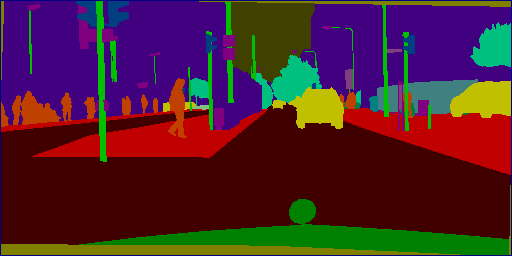} 
        \includegraphics[width=0.15\textwidth]{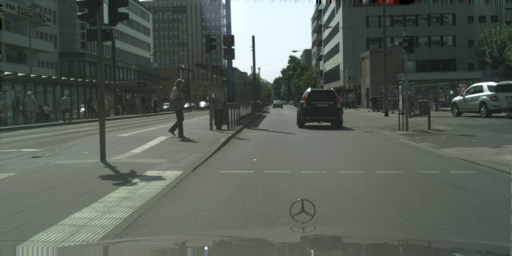} 
        \includegraphics[width=0.15\textwidth]{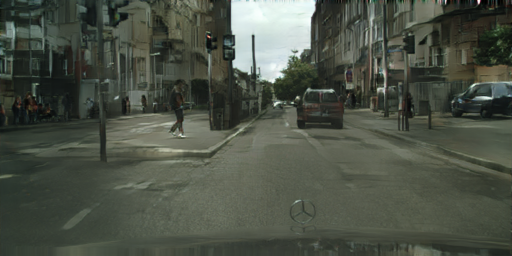} 
        \includegraphics[width=0.15\textwidth]{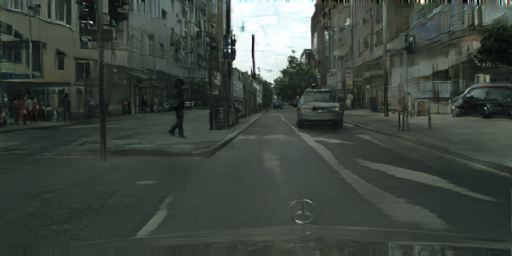} 
        \includegraphics[width=0.15\textwidth]{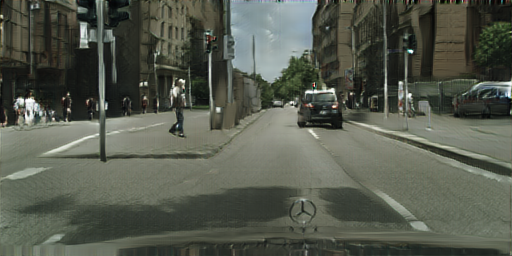} 
        \includegraphics[width=0.15\textwidth]{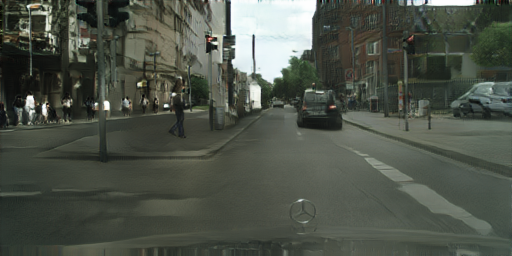} \\
        
        \includegraphics[width=0.15\textwidth]{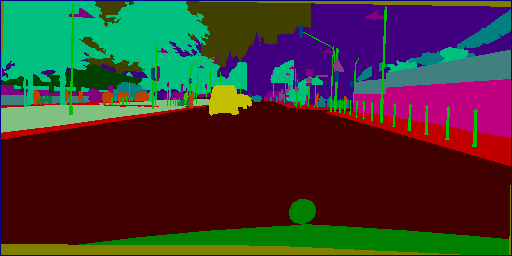} 
        \includegraphics[width=0.15\textwidth]{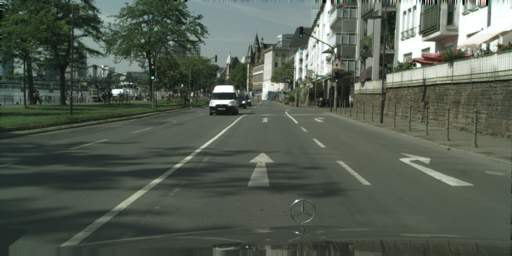} 
        \includegraphics[width=0.15\textwidth]{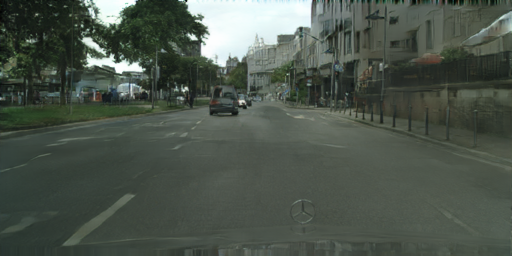} 
        \includegraphics[width=0.15\textwidth]{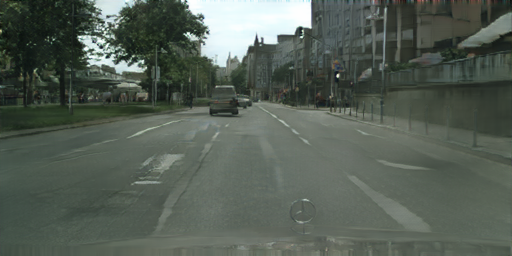} 
        \includegraphics[width=0.15\textwidth]{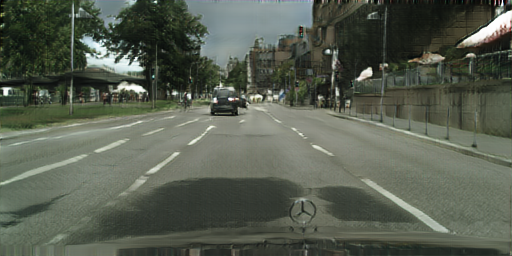} 
        \includegraphics[width=0.15\textwidth]{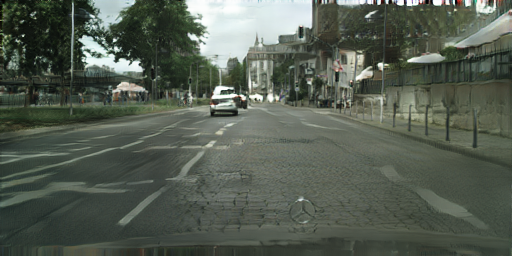} \\
        
        \includegraphics[width=0.15\textwidth]{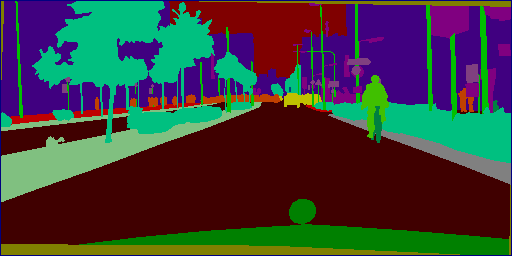} 
        \includegraphics[width=0.15\textwidth]{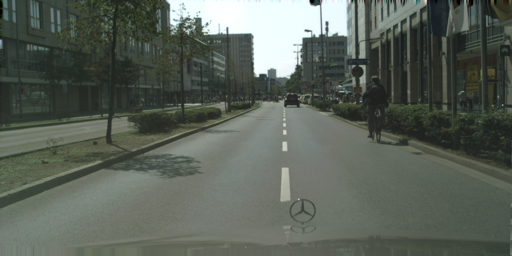} 
        \includegraphics[width=0.15\textwidth]{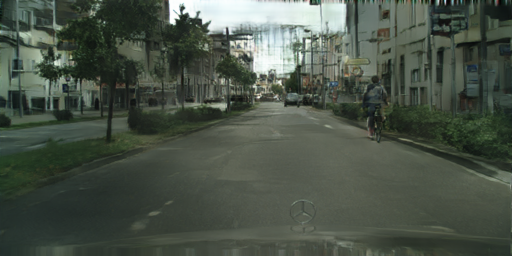} 
        \includegraphics[width=0.15\textwidth]{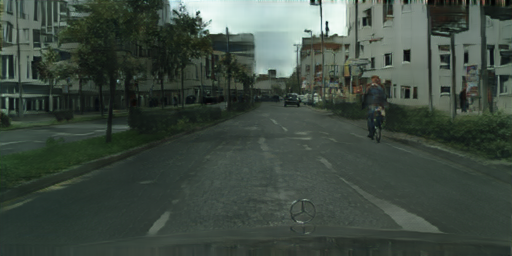} 
        \includegraphics[width=0.15\textwidth]{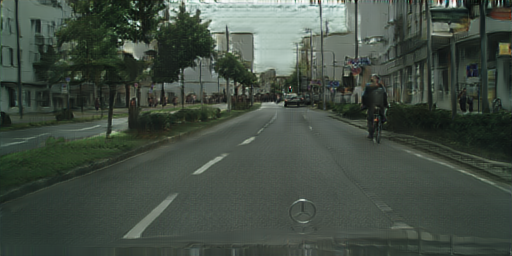} 
        \includegraphics[width=0.15\textwidth]{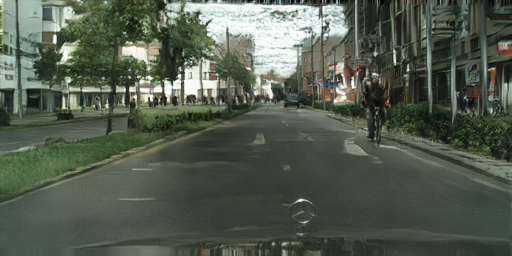} \\
        
        \includegraphics[width=0.15\textwidth]{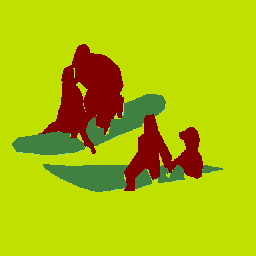} 
        \includegraphics[width=0.15\textwidth]{coco/gt/000000142971.jpg} 
        \includegraphics[width=0.15\textwidth]{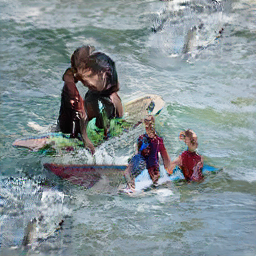} 
        \includegraphics[width=0.15\textwidth]{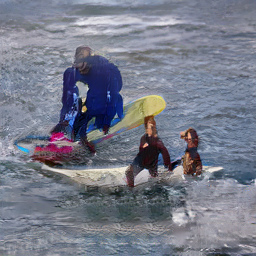} 
        \includegraphics[width=0.15\textwidth]{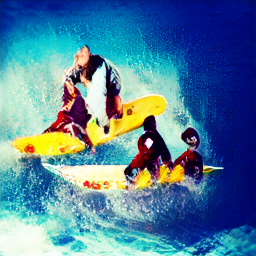} 
        \includegraphics[width=0.15\textwidth]{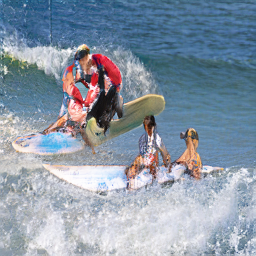} \\
        
        \includegraphics[width=0.15\textwidth]{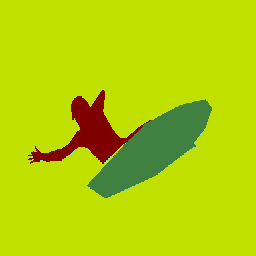} 
        \includegraphics[width=0.15\textwidth]{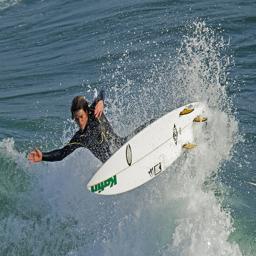} 
        \includegraphics[width=0.15\textwidth]{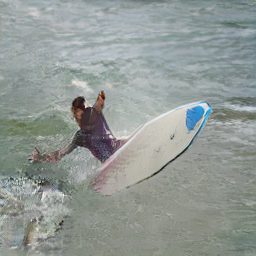} 
        \includegraphics[width=0.15\textwidth]{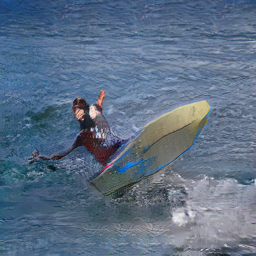} 
        \includegraphics[width=0.15\textwidth]{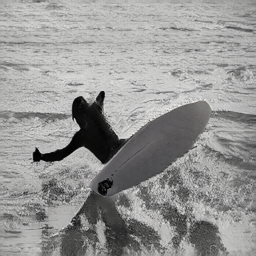} 
        \includegraphics[width=0.15\textwidth]{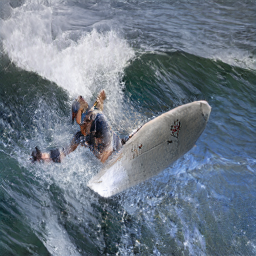} \\
        
        \includegraphics[width=0.15\textwidth]{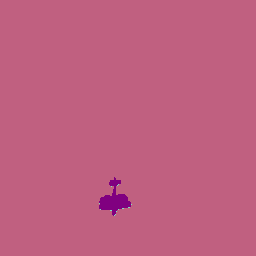} 
        \includegraphics[width=0.15\textwidth]{coco/gt/000000217060.jpg} 
        \includegraphics[width=0.15\textwidth]{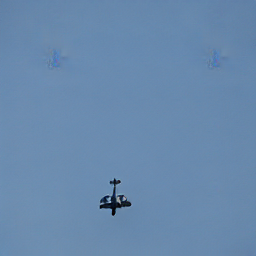} 
        \includegraphics[width=0.15\textwidth]{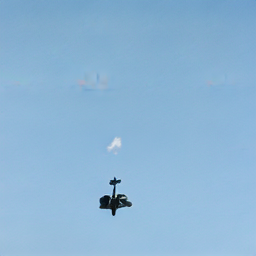} 
        \includegraphics[width=0.15\textwidth]{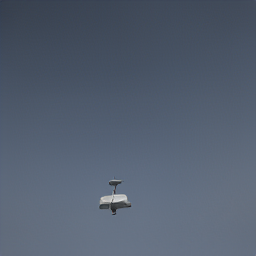} 
        \includegraphics[width=0.15\textwidth]{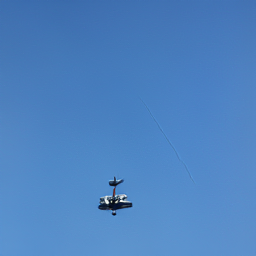} \\
        
        \includegraphics[width=0.15\textwidth]{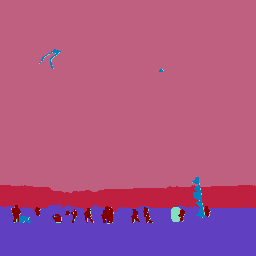} 
        \includegraphics[width=0.15\textwidth]{coco/gt/000000305317.jpg} 
        \includegraphics[width=0.15\textwidth]{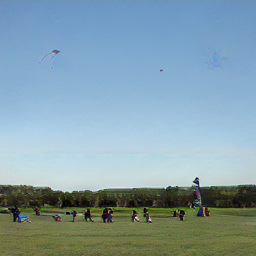} 
        \includegraphics[width=0.15\textwidth]{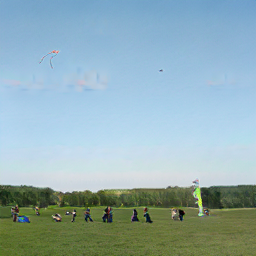} 
        \includegraphics[width=0.15\textwidth]{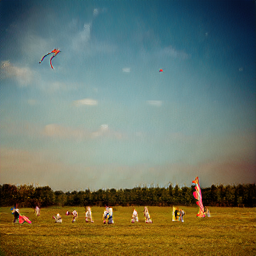} 
        \includegraphics[width=0.15\textwidth]{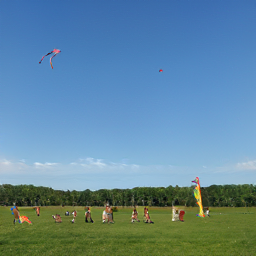} \\
        
    \caption{Generated samples in Cityscapes and COCO-Stuff. From the left to the right are input semantic layout, ground truth, SAPDE, CLADE-ICPE, OASIS, and VA-OASIS (ours). The first three rows are from Cityscapes while the left images are from COCO-Stuff.}
    \label{fig:vis}
\end{figure*}

\textbf{More generated results.} Figure \ref{fig:vis} displays more generated images for Cityscapes and COCO-Stuff dataset. From the figure, OASIS results in same shape of shadow in the first two rows while SPADE and CLADE-ICPE lead to repeating artifacts in the water (fourth and fifth row) or sky (the last two rows). The generated images by OASIS look like not nature, such as the grey water and dark sky, though it does not output similar pattern. In contrast, our method leads to more natural images without repeating artifacts. Specially, the second row suggests that our algorithm obtains diverse contrast and illumination for grasses and trees in one image and thus owns better intra-class variation.

\subsection{Ablation study}

\begin{table*}[h]
\begin{center}
\begin{tabular}{|c|c|c|c|c|c|c|c|}
\hline
 & $\gamma_n$ & $\gamma_{p}$ & FID-v$\downarrow$ & mIoU$\uparrow$ & Acc$\uparrow$ & para & FLOPs \\
\hline
\multirow{4}{*}{VA-SPADE} 
& no & no & 57.22 & 62.66 & 93.49 & 93.05 & 281.64 \\
& yes & no & 54.11 & 61.78 & 93.38 & 81.22 & 189.37 \\
& no & yes & 54.92 & 61.64 & 93.51 & 95.21 & 282.12 \\
& yes & yes & 53.30 & 61.08 & 93.40 & 83.34 & 189.61 \\
\hline
\multirow{4}{*}{VA-CLADE}
& no & no & 57.04 & 57.65 & 92.94 & 67.90 & 75.58 \\
& yes & no & 55.58 & 61.89 & 93.42 & 68.29 & 75.58 \\
& no & yes & 58.52 & 61.64 & 93.57 & 70.06 & 76.06 \\
& yes & yes & 55.18 & 62.25 & 93.45 & 70.41 & 75.82 \\
\hline
\end{tabular}
\end{center}
\caption{Ablation study results of VA-SPADE and VA-CLADE on Cityscapes \cite{cordts2016cityscapes}. As the discriminator is same in all cases, only the characters of generator is given.}
\label{table:ablation}
\end{table*}

\begin{table*}[h]
\begin{center}
\begin{tabular}{|c|c|c|c|c|c|c|c|}
\hline
 & semantic-noise & position-code & FID-v$\downarrow$ & mIoU$\uparrow$ & Acc$\uparrow$ & para & FLOPs \\
\hline
\multirow{6}{*}{VA-SPADE} 
& $\gamma_n$ & $p_l$ & 53.30 & 61.08 & 93.40 & 83.34 & 189.61 \\
& one-channel & $p_l$ & 54.21 & 60.46 & 93.23 & 83.31 & 189.38 \\
& plus & $p_l$ & 52.70 & 60.66 & 93.41 & 96.78 & 282.12 \\
& rand & $p_l$ & 53.54 & 60.35 & 93.16 & 83.34 & 189.61 \\
& $\gamma_n$ & $p_a$ & 54.09 & 60.46 & 93.26 & 81.25 & 189.61\\
& $\gamma_n$ & $p_r$ & 54.14 & 60.86 & 93.43 & 81.25 & 189.61\\
\hline
\multirow{6}{*}{VA-CLADE} 
& $\gamma_n$ & $p_l$ & 55.18 & 62.25 & 93.45 & 70.41 & 75.82 \\
& one-channel & $p_l$ & 56.14 & 61.56 & 93.50 & 70.38 & 75.58 \\
& plus & $p_l$ & 53.17 & 61.20 & 93.32 & 71.59 & 76.06 \\
& rand & $p_l$ & 55.54 & 59.83 & 93.09 & 70.41 & 75.82 \\
& $\gamma_n$ & $p_a$ & 54.54 & 59.84 & 93.32 & 68.32 & 75.82 \\
& $\gamma_n$ & $p_r$ & 55.64 & 61.99 & 93.56 & 68.32 & 75.82 \\
\hline
\end{tabular}
\end{center}
\caption{Variants of semantic noise and position code in VA-SPADE and VA-CLADE on Cityscapes. As the discriminator is same in all cases, only the situation of generator is given.}
\label{table:variant}
\end{table*}

We design an ablation study to understand the contribution from semantic noise and position code. Table \ref{table:ablation} displays the experimental results on Cityscapes dataset. From the table, both them benefit FID-v but harm mIoU and Acc for SPADE while position code alone gives adverse impact. We believe that it results from the non-overlap for semantic layout in the normalization layer of CLADE. In contrast, the vicinity is already considered in SPADE via the convolution operation with three as the kernel size. Further, semantic noise shrinks the number of parameters and computations while position code increase both of them. Combining them together tends to contribute FID, as well as mIoU and ACC for CLADE. In terms of SPADE, the combination only contributes to FID but is inferior in mIoU and Acc mainly because of the exist of the convolution operation. To shortly conclude, context information is useful for semantic image synthesis, as well as intra-class variation.

\subsection{Variants}

Except for semantic noise and learnable position code, we aim to analyze their variants. For the semantic noise, we consider the number of channels with $\gamma_s \in \mathbb{R}^{B\times 1 \times H' \times W'}$, inspired by the relative position code in CLADE. But we believe that channel one results in worse performance because of its representation capacity. Further, the concatenation to combine noise $\gamma_{n}$ and semantic layout feature $\gamma_{s}$ can be replaced by element-wise plus with which the number of channels should be $C'$, instead of $\frac{C'}{2}$. Moreover, random noise replaces with semantic noise. The three cases are denoted as one-channel, plus, and rand. For position code, learnable version $p_l$ can be replaced with absolute one $p_a$ and relative one $p_r$, as mentioned before. Similarly, the number of channel can also set as one. The experimental results are displayed in Table \ref{table:variant}.

From the table, we can see that one-channel slightly produces worse performance for both VA-SPADE and VA-CLADE with smaller decreases in number of parameters and computation. Element-wise plus of semantic noise $\gamma_n$ and semantic layout feature $\gamma_s$ benefits FID-v but cost at huge number of parameters and FLOPs, as well as lower mIoU and Acc, which suggests that balance the computation and performance is still a challenge in our algorithms. Besides, replacing semantic noise with random noise always outputs higher FID-v and a lower mIoU and Acc. Based on this situation, we argue that random noise contributes to intra-class variation but may lead to lower inter-class variation while our semantic noise achieves a balance. Finally, absolute and relative position code result in a little loss in all of the three evaluations with a less number of parameters.

\section{Conclusion}
\label{sec:conclusion}

In this paper, we highlighted the class-level mode collapse in semantic image synthesis, the image-level mode collapse is heavily discussed in the literature though. We gave a deep analysis to show the phenomenon and reason of the collapse in three current state-of-the-art algorithms, SPADE, CLADE, and OASIS. Beside, inter- and intra-class variation are defined to understand the class-level mode collapse. We found that the conditional normalization layer in the current algorithms contributes to inter-class variation but the intra-class variation is not enough. Further, we introduce two simple mechanisms, semantic noise and learnable position code, to increase the intra-class variation. The extensive experimental results suggest that our method is beneficial to semantic image synthesis with better performance with same training conditions than current state-of-the-art. In spite of success, the performance of our method is still not visually applicable, compared to natural images, and we only focus on the architecture of generator. We leave it as our future work to consider the semantic image synthesis in a systematic way, including framework of generator and discriminator, along with loss functions.

{\small
\bibliographystyle{ieee_fullname}
\bibliography{11_references}
}

\ifarxiv \clearpage \appendix
\label{sec:appendix}

% \section{}
 \fi

\end{document}